\documentclass[lettersize,journal,twoside]{IEEEtran}
\usepackage{amsmath,amsfonts}
\usepackage{array}
\usepackage{textcomp}
\usepackage{stfloats}
\usepackage{url}
\usepackage{graphicx}
\usepackage{cite}
\usepackage[colorlinks=true, allcolors=blue]{hyperref}
\usepackage[ruled, lined, longend, linesnumbered]{algorithm2e}
\usepackage{flushend}
\begin{document}

\title{
TrustChain: A Blockchain Framework for Auditing and Verifying Aggregators in Decentralized Federated Learning}

\author{Ehsan~Hallaji,~\IEEEmembership{Member,~IEEE,}
		Roozbeh~Razavi-Far,~\IEEEmembership{Senior Member,~IEEE,}
		Mehrdad~Saif,~\IEEEmembership{Fellow,~IEEE}

   \thanks{This work was supported by the Natural Sciences and Engineering Research Council of Canada (NSERC) under funding reference numbers CGSD3-569341-2022 and  RGPIN-2021-02968.}
	\thanks{Ehsan Hallaji and Mehrdad Saif are with the Department of Electrical and Computer Engineering, University of Windsor, Windsor, ON N9B 3P4, Canada. E-mail: hallaji@uwindsor.ca, msaif@uwindsor.ca.\protect}
	\thanks{Roozbeh Razavi-Far is with the Faculty of Computer Science, University of New Brunswick, Fredericton, NB E3B 5A3, Canada, and also with the Department of Electrical and Computer Engineering, University of Windsor, Windsor, ON N9B 3P4, Canada. E-mail: roozbeh.razavi-far@unb.ca.\protect}}

\markboth{IEEE Transactions on Big Data, November 2025}%
{Hallaji \MakeLowercase{et al.}: TrustChain: A Blockchain Framework for Auditing and Verifying Aggregators in DFL}

\maketitle

\begin{abstract}
The serverless nature of Decentralized Federated Learning (DFL) requires allocating the aggregation role to specific participants in each federated round. Current DFL architectures ensure the trustworthiness of the aggregator node upon selection. However, most of these studies overlook the possibility that the aggregating node may turn rogue and act maliciously after being nominated. To address this problem, this paper proposes a DFL structure, called \texttt{TrustChain}, that scores the aggregators before selection based on their past behavior and additionally audits them after the aggregation. To do this, the statistical independence between the client updates and the aggregated model is continuously monitored using the Hilbert-Schmidt Independence Criterion (HSIC). The proposed method relies on several principles, including blockchain, anomaly detection, and concept drift analysis. The designed structure is evaluated on several federated datasets and attack scenarios with different numbers of Byzantine nodes.
\end{abstract}


\section{Introduction}
\label{sec:intro}
\IEEEPARstart{T}{he} advent of Federated Learning (FL) advanced the field of distributed machine learning by introducing data decentralization as a solution to bring about data privacy and communication efficiency \cite{pmlr-v54-mcmahan17a}. Despite its advantages, FL was shown to be vulnerable against a spectrum of adversaries due to its distributed nature \cite{pmlr-v97-bhagoji19a, NEURIPS2020_b8ffa41d}. Numerous research endeavors have been dedicated to studying these threats and finding robust defense mechanisms to mitigate them. 

A common perception among the majority of these studies is that the server is trustworthy, and malicious activities can potentially be initiated from the edge nodes. Thus, available defense mechanisms are mostly concerned with client activities within the FL network. Another issue of concern in FL is the fact that the server could be a single point of failure \cite{Yang_Ghaderi_2024, Xiong_Yan_Wang_Li_2024}. From a security standpoint, this can translate into the vulnerability of the FL network when the server is breached or when the server itself is not trustworthy.

These security concerns motivated researchers to come up with decentralized architectures of FL that do not rely on a central server \cite{9292450}, which in turn improves the reliability and scalability of FL. In this structure, the client-side operations are similar to that of FL. However, the aggregation role is performed by one or more sets of trusted nodes. Several approaches are available for selecting the aggregator nodes in each round. While some works follow approaches as simple as following a random or specific order in selecting the aggregator nodes, others use more sophisticated technology such as Smart Contracts (SC) and proofs in blockchain to select aggregator nodes more confidently \cite{9321132, 9284684, 9293091}.

In the context of cyber security, several works are dedicated to securing DFL against privacy and performance attacks \cite{10420449}. To mitigate privacy attacks, these structures are often combined with homomorphic encryption, differential privacy, or secure multiparty communications to secure DFL communications and minimize the risk of inference attacks \cite{9777682, 10.1145/3426474}. Dealing with performance-related attacks such as poisoning attacks, on the other hand, requires in-depth analysis of generated updates using anomaly detection and robust aggregation protocols \cite{Yang_Ghaderi_2024, DBLP:journals/corr/abs-2202-02817}. Techniques used in both of these domains are mainly inspired by previous studies in FL, as privacy-preserving and client-update evaluation are the two main security concerns in FL.

Verifying the aggregator nodes, on the other hand, is less studied in the literature. Most works in this domain assume that once a trustworthy node is selected to carry out the aggregation, it will not become rogue. This contradicts the fact that Byzantine nodes initially act honestly in the network and become dishonest once they find the opportunity to execute a malicious action. In fact, to the best of our knowledge, only a limited number of research works target this problem \cite{ijcai2022p792}. Available solutions mostly use a trusted committee for verifying the aggregated model \cite{9321132}. Nevertheless, the trustworthiness of verifying nodes cannot be guaranteed.

To address this gap, we propose a novel approach, called \texttt{TrustChain}, that leverages blockchain technology to ensure the continuous trustworthiness of the aggregator node throughout the DFL process. This method consists of two main components:
\begin{itemize}
    \item Pre-Selection Evaluation (PSE): When miners request downloading the blocks for aggregation, an SC is triggered to calculate a score for participating miners based on the cosine similarity of the parameters they uploaded to the blockchain and the resulting aggregated model in each round, for a window of past DFL iterations. This score indicates the tendency of each node to drift from the training path of the global model. This step ensures that only nodes with legitimate and consistent updates are considered for the aggregation role, thereby minimizing the risk of selecting a potentially malicious node.
	\item Post-Aggregation Auditing (PAA): When the selected node requests to push the aggregated model to the blockchain, another SC is activated to measure the statistical independence between the aggregated model and the median of user updates using HSIC. To determine whether the obtained HSIC is safe or not, a threshold is dynamically set using the HSIC values included in previous blocks. If the new block is verified, the third SC is triggered and adds the estimated HSIC in this round to the block to facilitate the threshold estimation in the next round. This audit trail enables the detection of any malicious actions performed by the aggregator node during the aggregation process.
\end{itemize}

By integrating these two components, the proposed structure aims to enhance the security and robustness of DFL systems. The use of blockchain ensures that all transactions and updates are securely recorded, providing an additional layer of security and trust. Anomaly detection and concept drift analysis help in identifying suspicious behaviors and further strengthen the defense against malicious activities.

\section{Background}
\label{sec:background}
The literature review on the security aspects of DFL is divided into studies that analyze security threats in DFL and those designing defense mechanisms to safeguard DFL against adversaries. This section initially explains the concept of DFL and then overviews threats and defense mechanisms in this domain.

\subsection{Decentralized Federated Learning}
DFL primarily addresses the issue of the single point of failure in FL, by removing the dependency on a reliable server to carry out the aggregation and enhance communication efficiency \cite{9084352, MAL-083}. On the client side, the process of training the local models and preparing the updates for the global aggregator is similar to that of FL. However, due to the lack of a central server in DFL, exchanging model parameters and model aggregation are performed using peer-to-peer communications or blockchain technology \cite{10420449}. Integration with blockchain provides additional benefits such as traceability and immutability \cite{10535235,9705057}. Despite DFL's security and efficiency characteristics, the design is not faultless. For example, adding blockchain into FL may leave the system vulnerable to blockchain-related security threats. More importantly, the trustworthiness of participants is more critical in DFL, as the aggregating nodes are selected among them in each round.

\subsection{Threats}
Similar to FL, threats to DFL generally fall under privacy and performance-related attacks \cite{10420449}. Privacy attacks primarily aim at stealing sensitive data during inference based on the communicated gradients. An example of privacy attacks in DFL is presented by \cite{DBLP:journals/corr/abs-2202-02817}. On the other hand, poisoning attacks target the global model by injecting poisonous updates during the training. The attacker can use data and model poisoning to deteriorate the overall performance of the global model. For instance, \cite{9292450} performs data poisoning to disrupt the global performance. This study assumes 30 percent of the nodes are malicious, and considers that this ratio remains unchanged in the experiments. The same objective can be achieved using Gradient Manipulation (GM), in which Byzantine data holders generate forged gradients sampled from a Gaussian distribution \cite{8622598}. The same malicious objective is satisfied in \cite{8994206} by filliping the class labels randomly on the training data. Moreover, DFL can be exposed to more subtle attacks such as backdoors as presented in \cite{DBLP:journals/corr/abs-2202-02817, 10.1145/3422337.3447837}.

\subsection{Defenses}
The majority of studies on DFL security integrate blockchain with this structure to facilitate auditing the updates and participants. Nevertheless, some studies rely solely on peer-to-peer communications and do not employ blockchain \cite{DBLP:journals/corr/abs-1901-11173, roy2019braintorrent}. These works are primarily focused on eliminating the single point of failure in conventional FL. For instance, \cite{roy2019braintorrent} resorts to version control to perform server selection using an arbitrary pattern.

Biscotti was among the first efforts to secure DFL against privacy attacks through blockchain integration \cite{9292450}. The method integrates Differential Privacy (DP) with secure aggregation as a defense against poisoning and inference attacks. Another study proposes LearningChain, a DFL structure that also uses DP and an aggregation mechanism that works based on decentralized Stochastic Gradient Descent (SGD) to eliminate Byzantine attacks \cite{8622598}. One of the first efforts to demonstrate the use of smart contracts in DFL was made by \cite{9284684}. This method uses smart contracts to keep track of client states and store the global model. Proof of correctness can also be used by registered users to audit the generated updates \cite{8905038}. The sharding approach commonly used in blockchain is used by \cite{9210138} to secure the aggregation process. \cite{9321132} proposed VFChain, a structure that audits trainers and generates updates using a trusted committee. The committee members are selected arbitrarily, and they use a verifying contract to verify aggregated models. Another approach proposed by \cite{DBLP:journals/corr/abs-2202-02817} integrates several methods such as DP, anomaly detection, and gradient pruning to safeguard DFL against poisoning attacks. Zero-Knowledge (ZK) proofs is used in \cite{10535217} so that each round’s aggregation is verifiably faithful. To do so, the aggregator generates a proof of correct model aggregation, which blockchain miners validate without exposing client models, ensuring both integrity and privacy. VerifBFL extends this concept with ZK Succinct Non-interactive Arguments of Knowledge (ZK-SNARKs) and incrementally verifiable computation, allowing proofs to be generated for both local training and aggregation steps, which are then verified on-chain in under a second, offering trustless correctness but at the cost of cryptographic overhead \cite{11073628}. Secure Verifiable Aggregation frameworks combine public verifiable secret sharing with masking and encryption to hide individual updates while still enabling blockchain participants to verify the correctness of the aggregated model, providing dropout resilience in addition to integrity guarantees \cite{ZHU2022100046}. VDFChain employs verifiable delay functions and polynomial commitments to make aggregated updates verifiable while deterring manipulation \cite{ZHOU2024103814}. Committee-based consensus designs, such as BFLC, assign aggregation to a selected committee whose results are recorded on-chain, reducing reliance on a single aggregator and mitigating certain attacks \cite{9293091}. Verifiable off-chain computation frameworks also use ZK-proof systems to execute aggregation off-chain while producing proofs that are verified on-chain, reducing latency while maintaining integrity \cite{9881809}. BFL-SA focuses on enhancing secure aggregation in blockchain-based FL, protecting the integrity and confidentiality of aggregated updates with lightweight cryptographic protocols \cite{LIU2024103163}. To secure DFL in IoT, \cite{ABDMEZIEM2024101276} designs a hybrid node selection mechanism that combines reputation scores with reinforcement learning, and a multi-level aggregation scheme to improve reliability and handle asynchronicity. While this indirectly enhances aggregator trust by prioritizing capable nodes, their approach does not verify aggregator behavior after aggregation. 

\subsection{Knowledge Gap}
While recent studies have explored various blockchain-based mechanisms to enhance the security of DFL, most existing solutions secure the aggregator either through cryptographic verifiability or trusted node selection. Cryptographic proof-based methods ensure that aggregation is performed correctly, but they incur significant computational and communication overhead, require specialized proof systems, and may be challenging to scale in large, resource-constrained deployments. Trusted selection approaches indirectly improve aggregator trustworthiness by reducing the likelihood of selecting a malicious node. However, these methods do not verify aggregator behavior after aggregation and thus cannot detect malicious actions from otherwise trusted aggregators. Furthermore, several works focus on complementary issues, such as privacy preservation, dropout resilience, and asynchronicity handling, without providing continuous, fine-grained verification of aggregation integrity throughout training rounds.

This reveals a clear gap: there is a lack of lightweight, topology-agnostic mechanisms that can continuously monitor and verify the trustworthiness of the aggregator’s behavior during the learning process, without relying on heavy cryptography or fixed committee structures. \texttt{TrustChain} addresses this gap through a behavior-based trust framework that is adaptable to various FL settings and complementary to proof-based methods. Its PSE filters aggregators before selection using historical performance patterns, while its PAA statistically analyzes aggregated models for anomalies after aggregation. This combination provides continuous, low-overhead security that can operate independently or be integrated with existing blockchain-based FL frameworks for layered protection.

\section{Problem Formulation}
\label{sec:problem}
Given a set of $n$ client nodes in the DFL network $C=\{c_1,c_2,\dots,c_n\}$, each node $c_i$ trains a local model $M_i$, and uploads its parameters $\theta_i$ to the blockchain $B$. The blockchain is formulated as $B=\{b^1, b^2, \dots, b^t\}$, where $b_t$ denotes the last block at round $t$ of DFL, and the others indicate the previous blocks in sequence. In this formulation, $t$ is dynamic as the size of $B$ changes in time.

In each round $t$, the last $r$ number of blocks that contain $\theta_i^t$ should be aggregated to update the global state of the model $\Bar{\theta}^t$. Depending on the DFL structure, one node $c^*$ will be selected to carry out the aggregation task, which is also referred to as the selected miner of the blockchain network. Normally, $c_a$ aggregates $\{\theta^t_i\}_{i=1}^r$ into $\bar{\theta}^t$ and uploads it to the blockchain as $b^t \leftarrow \bar{\theta}^t$. 

In this scenario, the two main attack surfaces are the client updates and the aggregated parameters. Firstly, $\theta_i$ uploaded to $B$ may be poisoned to satisfy a malicious goal. These threats are common with the centralized architecture of FL and are commonly addressed using defense mechanisms such as robust aggregation and anomaly detection. Secondly, even if the updates are filtered and the malicious updates are dropped, the node aggregating the results $c^*$ can still compromise the entire network by updating the global state into a malicious set of parameters.

\section{Threat Models}
\label{sec:threat}

While the spectrum of poisoning attacks is vast, this paper focuses on model poisoning attacks. We consider four poisoning attacks, namely label flipping, sign flipping, GM, and Training Objective Manipulation (TOM). These attacks were selected because they represent both simple-yet-effective poisoning strategies (e.g., label/sign flipping) and more sophisticated attacks (e.g., GM, TOM) that can evade conventional defenses. For each attack, we vary the proportion of Byzantine nodes in the population (1–20\%) to study performance under different threat intensities. Attacks are injected in two distinct ways: (i) Client-side attackers, where malicious nodes contribute poisoned local updates during training, and (ii) Aggregator-side attackers, where the selected aggregator is malicious and directly uploads a poisoned global model, bypassing the robust aggregation stage. The latter case is critical for evaluating \texttt{TrustChain}, as it explicitly targets our main security objective: detecting and mitigating rogue aggregators.

When these attacks are initiated from the client side, a robust aggregator attempts to mitigate their effects. If a malicious node is selected as an aggregator, it can bypass the robust aggregator by directly feeding the poisoned parameters to the DFL. These attacks are formally explained as follows.

\paragraph{Label flipping} The labels of the training data are flipped to obtain corrupted parameters \cite{247652, HALLAJI2023110384}. Let $\mathcal{D}_i = {(x_i, y_i)}$ be the dataset of node $c_i$, where $y_i$ is the true label. The attacker generates a poisoned dataset $\tilde{\mathcal{D}}_i$ by randomly flipping labels:
\begin{equation}
\tilde{\mathcal{D}}_i = \{(x_i, \hat{y}_i) \mid P(\hat{y}i \neq y_i) = \gamma\}_{i=1}^m,
\end{equation}
where $\gamma$ is the proportion of flipped labels, and $m$ is the number of training samples. Poisoned parameters $\tilde{\theta}_i$ used for attacking the DFL system are derived from this corrupted dataset:
\begin{equation}
\tilde{\theta}_i \gets M_i(\tilde{\mathcal{D}}_i).
\end{equation}

\paragraph{Sign flipping} The attacker reverses the sign of the gradients during parameter estimation \cite{Li_Xu_Chen_Giannakis_Ling_2019}. For a gradient from node $c_i$, the poisoned gradient is the negative of the estimated gradients. The corresponding poisoned parameters are:
\begin{equation}
\tilde{\theta}_i = \theta_i - \eta \cdot \text{sign}\big(\nabla \mathcal{L}(\theta_i)\big),
\end{equation}
where $\eta$ is the learning rate, $\theta$ is the set of network parameters, and $\mathcal{L}$ denotes the loss function. 

\paragraph{Gradient manipulation} During the aggregation phase, the attacker introduces noise into the aggregated parameters \cite{pmlr-v139-li21h}. Let $\{\theta_i^t\}_{i=1}^r$ represent the set of parameters from the last $r$ blocks. The attacker modifies the aggregated update as:
\begin{equation}
\tilde{\theta}^t = \frac{1}{r} \sum_{i=1}^r \theta_i^t + \Delta \theta,
\end{equation}
where $\Delta \theta$ is the noise injected by the attacker.

\paragraph{Training objective manipulation} The attacker modifies the training objective by adding a penalty term that deteriorates the classification performance \cite{pmlr-v97-bhagoji19a}. If $\mathcal{L}(\theta)$ is the original loss function, the attacker changes it to:
\begin{equation}
\tilde{\mathcal{L}}(\bar{\theta}^t) = \mathcal{L}(\bar{\theta}^t) + \zeta \cdot \mathcal{P}(\bar{\theta}^t),
\end{equation}
where $\mathcal{P}(\bar{\theta}^t)$ is a penalty term added by the attacker, and $\zeta$ scales the severity of the attack. We define the penalty term as:
\begin{equation}
    \mathcal{P}(\bar{\theta}^t) = \|\nabla \mathcal{L}(\bar{\theta}^t) - \nabla \mathcal{L}_{target} \|^2,
\end{equation}
where $\mathcal{L}_{target}$ is the direction towards which the gradients are guided. Here, we choose $\mathcal{L}_{target} = - \nabla \mathcal{L}(\bar{\theta}^t)$, which results in $\mathcal{P}(\bar{\theta}^t) = \|2 \cdot \nabla \mathcal{L}(\bar{\theta}^t)\|^2$.

\section{Blockchain-Enabled Trustworthy Aggregation}
\label{sec:design}

If DFL operates without an aggregator evaluation mechanism, aggregator nodes may be chosen using methods that do not account for their trustworthiness, which increases the risk of selecting a malicious node. In the proposed method, aggregator security is ensured through PSE and PAA mechanisms. PSE operates before selection and scores miners over a sliding window of $q$ previous rounds based on their historical alignment with the global model’s trajectory using concept drift analysis. This allows the system to prioritize the most trustworthy candidate and significantly reduce the likelihood of a rogue node being selected. PAA operates after aggregation and statistically examines the aggregated model to identify anomalies or signs of poisoning. If PSE is missing, the selection process lacks behavior-based filtering, increasing the chance that an untrustworthy aggregator is chosen. If PAA is missing, there is no safeguard to detect an aggregator that behaves maliciously after being selected, which allows a compromised node to poison the model without detection. Together, PSE and PAA form a layered defense where PSE filters aggregators before selection and PAA verifies their behavior afterward.

Given that the goal of this study is to ensure the trustworthiness of the aggregator, the proposed \texttt{TrustChain} assumes the employed robust aggregation algorithm is reliable. Consequently, this research focuses on securing the aggregation phase by performing PSE and PAA in DFL. The process begins with the PSE algorithm evaluating the past behavior of participants using concept drift analysis to determine a set of trustworthy candidates. Then, the aggregated model is further evaluated by PAA based on the statistical independence of $\bar{\theta}^t$ with that of the previous block and $\theta_i$ received in the current round. The following subsections describe each component in detail, with the final subsection dedicated to the complexity analysis.

\subsection{Robust Aggregation}
In our experiments, we use the \texttt{TrimmedMean} algorithm as the robust aggregator \cite{pmlr-v80-yin18a}. Robust aggregation aims to mitigate the influence of poisoned or extreme client updates by limiting the effect of outliers in each model parameter coordinate. Let $\{\theta_i\}_{i=1}^r$ denote the set of full model parameter vectors received from $r$ clients in a round, where each $\theta_i$ is trained locally on $m$ samples. For each parameter coordinate $j$, \texttt{TrimmedMean} collects scalar values $\theta_i^{j}$ across all clients, sorts them, removes a fixed fraction of the largest and smallest values, and averages the remaining ones. This coordinate-wise trimming is applied independently for all dimensions of the model parameter vector, producing the aggregated model $\bar{\theta}$.

The choice of robust aggregator is not part of \texttt{TrustChain}’s novelty. \texttt{TrimmedMean} is used here solely as a representative baseline due to its simplicity and established Byzantine robustness, ensuring a fair and reproducible evaluation of \texttt{TrustChain}. However, more advanced robust aggregators can be substituted without altering the operation of PSE or PAA.

\subsection{Pre-Selection Evaluation}

\begin{algorithm}[t]
\SetAlgoLined
\KwIn{Blockchain $B$, miner nodes $C^*$, number of recent blocks $q$, decay rate $\alpha$}
\KwOut{Selected aggregator node $c_a$}
\SetKwInput{Initialization}{Initialization}
\For{$1\leq i \leq q$}{
    Initialize score $S_i^t = 0$
}
\For{$\forall b_j \in \{b_j \mid b_j\in B\}_{j=t-q+1}^t$}{
    Retrieve the global state $\bar{\theta}^{j} \gets b_j$\\
    \For{$1 \leq i \leq q$}{
        Retrieve model parameters $\theta_i^{j-1}$ from $B$\\
        Calculate $\operatorname{CosSim}(\theta_i^{j-1}, \bar{\theta}^{j})$\\
        Update $S_i^t \gets S_i^t + \alpha^{t-j} \cdot \operatorname{CosSim}(\theta_i^j, \bar{\theta}^{j+1})$\\
    }
}
Select aggregator $c_a \gets \arg\max_{c^*_i \in C^*} S_i^t$\\
\Return $c_a$
\caption{Pre-Selection Evaluation}
\label{alg:selection}
\end{algorithm}

\begin{figure*}[t]
    \centering
    \includegraphics[trim={0.6cm 0.5cm 0.9cm 0.5cm},clip,width=\textwidth]{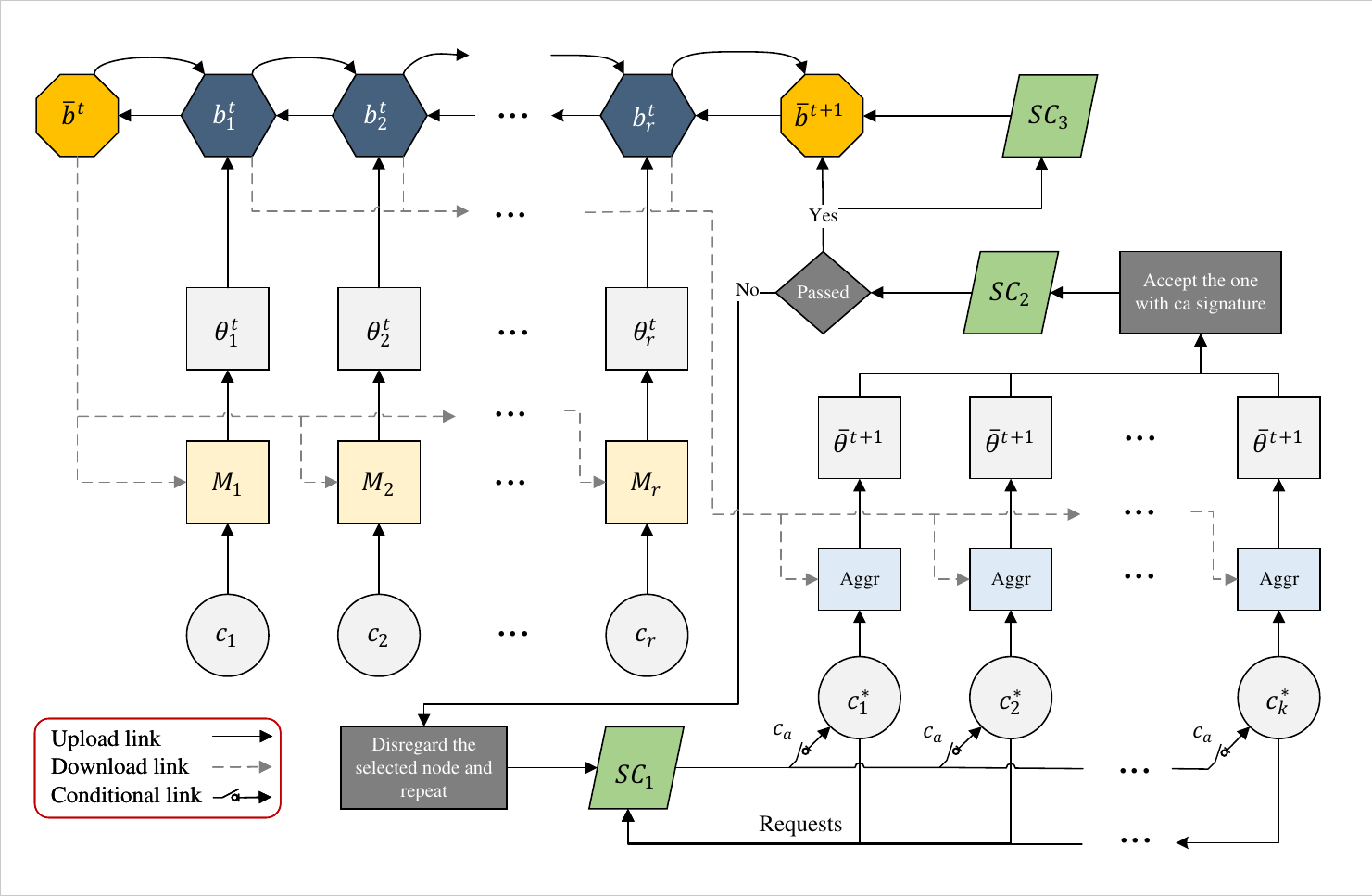}
    \caption{Block diagram of the proposed \texttt{TrustChain} structure for one DFL round at time $t$. Aggr denotes the robust aggregation algorithm. For simplicity, the diagram assumes each block number $i$ corresponds to node $c_i^t$. In this formulation, $t$ indicates the DFL iteration number. In addition, $b$ and $\bar{b}$ refer to blocks containing user updates $\theta$ and aggregated models $\bar{\theta}$, respectively.}
    \label{fig:diagram}
\end{figure*}

In order to determine the trustworthiness of nodes for the aggregation phase, they are scored based on their past behavior. Specifically, we monitor the tendency of $c_i$ to deviate from the $\bar{\theta}$ path in terms of direction and magnitude. As detailed in Algorithm \ref{alg:selection}, this goal is achieved by measuring the cosine similarity of updates $\theta_i^{t-1}$ with the corresponding aggregated model for which they are used for $\bar{\theta}^{t}$. 
\begin{equation}
    \operatorname{CosSim}(\theta_i^{t-1}, \bar{\theta}^t)=\frac{\theta_i^{t-1} \cdot \bar{\theta}^t}{\|\theta_i^{t-1}\| \|\bar{\theta}^t\|},
\end{equation}
where $1\leq i \leq r$. Note that we estimate the similarity for all nodes that participated in round $t-1$ regardless of the fact that not all $\theta_i^{t-1}$ may have been used to estimate $\bar{\theta^t}$. This approach results in a lower score for $c_i$ whose $\theta_i$ are disregarded or not used during the aggregation. To analyze the node behavior in a longer time span, a window of $q$ blocks is used to reveal the overall drift $c_i$ creates in the path of $\bar{\theta}$. Correspondingly, a score is calculated for each node at each time step.
\begin{equation}
S_i^t = \sum_{j=t-q+1}^{t} \alpha^{t-j} \cdot \operatorname{CosSim}(\theta_i^{j-1}, \bar{\theta}^{j}),
\end{equation}
where $\alpha$ is the rate of decay. The exponential decay used in this equation enables us to prioritize recent behavior and give less contribution weight to older scores. The aggregator is then selected from the pool of $k$ participating miners $C^*=\{c_1^*, c_2^*, \dots, c_k^*\}$ as:
\begin{equation}
    c_a = \arg\max_{c_i^*\in C^*}S_i^t
\end{equation}

PSE continuously monitors concept drift to ensure a candidate aggregator’s updates remain consistent with the evolving concept represented by the global model trajectory. Here, the concept refers to the statistical relationship between successive aggregated models and the local updates that contributed to them. For each node $c_i$, we track its historical behavior over the last $q$ rounds by measuring the cosine similarity between its submitted parameters $\theta_i^{t-1}$ and the corresponding aggregated model $\bar{\theta}^t$. This similarity captures both direction and magnitude alignment with the global model’s learning path. A sustained drop in similarity, especially when recent rounds are weighted more heavily via the decay factor $\alpha^{t-j}$, indicates a drift from the consensus learning trajectory, which may be due to malicious behavior (e.g., poisoning) or degraded participation quality. The aggregated score $S_i^t$ across the sliding window thus functions as a drift indicator so that high scores signal stable, aligned contributions, while low scores reveal significant concept drift. PSE uses this drift score to exclude candidates whose historical updates suggest a high likelihood of deviating from the global model’s objective, thereby reducing the risk of appointing a malicious or unreliable aggregator.

\subsection{Post-Aggregation Auditing}
Algorithm \ref{alg:validation} details the evaluation process for validating the aggregated model. Once $c_a$ pushes $\bar{\theta}^{t+1}$ to the blockchain, a SC is triggered to check the statistical independence of the aggregated model with parameters used in this round $\{\theta_i^t\}_{i=1}^r$ and the current state of the global model $\bar{\theta}^{t}$. In this process, statistical independence is measured using HSIC. Let $A$ and $B$ be two random variables. Then, HSIC between two variables can be formally defined as:
\begin{equation}
\begin{split}
&\operatorname{HSIC}(A, B) = \\
&~\mathbb{E}_{A B A^\prime B^\prime}[f_A(A,A^\prime) f_{B^\prime}(B, B^\prime)] \\
& + \mathbb{E}_{AA^\prime}[f_A(A,A^\prime)] \mathbb{E}_{BB^{\prime}}[f_B(B,B^\prime)] \\
& - 2\mathbb{E}_{AB}\big[\mathbb{E}_{A^\prime}[f_A(A, A^\prime)] \mathbb{E}_{Y^\prime}[f_B(B,B^\prime)]\big],    
\end{split}
\end{equation}
where $(\cdot)^\prime$ denotes independent copies of variables, and $f_{(\cdot)}$ indicates kernel functions. Efficient calculation of HSIC is often undertaken using an empirical estimate \cite{10.1007/11564089_7}:
\begin{equation}
    \operatorname{HSIC}(A, B) = \frac{1}{(|A|-1)^{2}}\operatorname{tr}(F_A H F_B H),
\end{equation}
where $|\cdot|$ returns the cardinality. In addition, $H=\mathbf{I} - \frac{1}{|\theta|}\mathbf{11}^\top$ is the centring matrix, and $F_A=f_A(a_i,a_j)$ and $F_B = f_B(b_i,b_j)$ are kernel matrices obtained from $a_i\in A$ and $b_i\in B$.

\begin{algorithm}[t]
\SetAlgoLined
\KwIn{Blockchain $B$, set of miner nodes $C^*$, aggregated model $\bar{\theta}^{t+1}$}
\KwOut{Validation result: approved or rejected}
Retrieve $\bar{\theta}^{t}$ from $b^{t}\in B$\;
Retrieve set of model parameters $\{\theta_i^t\}_{i=1}^r$ from  $B$\\
Calculate $\theta^t_m \gets \operatorname{Median}(\{\theta_i^t\}_{i=1}^r)$\\
Retrieve $h$ from previous $q$ blocks in $B$\\
Estimate $\tau^t$ using (\ref{eq:tau})\\
Calculate $\operatorname{HSIC}(\theta^t_m, \bar{\theta}^{t+1})$\;
\uIf{$\operatorname{HSIC}(\theta^t_m, \bar{\theta}^{t+1}) \geq \tau^t$}{
    Approve $\bar{\theta}^{t+1}$\\
    Trigger $SC_3$ to include HSIC in the next block\\
}
\Else{
    Reject $\bar{\theta}^{t+1}$\\
    Forbid  $c_a$ and reset the selection process\\
    Trigger $SC_1$ to select a new aggregator\\
}
\caption{Post-Aggregation Auditing}
\label{alg:validation}
\end{algorithm}

In order to use HSIC for measuring statistical independence of $\bar{\theta}^{t+1}$ with available $\theta_i^t$, we need to minimize the cost of this process, as calculating pairwise HSIC may become computationally burdensome for large values of $r$. Moreover, a threshold needs to be defined to determine whether the estimated HSIC is abnormal. Thus, we choose $\theta^t_m=\operatorname{Median}(\{\theta_i^t\}_{i=1}^r)$ as a representative of all $\theta_i^t$ and calculate $\operatorname{HSIC}(\theta_m^t,\bar{\theta^{t+1}})$ instead. Another alternative to using the median is calculating the average of $\{\theta_i^t\}_{i=1}^r$. However, the median is more robust against poisonous updates. At each round $t$, a threshold $\tau$ is obtained as:
\begin{equation}
    h = \bigcup_{t=1}^q \operatorname{HSIC}(\theta_m^t, \bar{\theta}^{t+1}),
    \label{eq:h}
\end{equation}
\begin{equation}
    \tau^t = \min(h) - \lambda\sigma,
    \label{eq:tau}
\end{equation}
where $h$ is the set of previous $q$ HSIC values, $\lambda$ is the scaling factor, and $\sigma$ is the standard deviation. $\bar{\theta}^{t+1}$ will be approved if $\operatorname{HSIC}(\theta_m^t, \bar{\theta}^{t+1}) \geq \tau^t$. Note that in (\ref{eq:h}), only $\operatorname{HSIC}(\theta_m^t, \bar{\theta}^{t+1})$ is estimated in time $t$, as the HSIC for previous time steps is already available in previous blocks.

PSE and PAA are seperated intentionally. PSE acts early to filter out clearly harmful updates before aggregation, reducing the computational burden and limiting the influence of severe outliers, while PAA operates afterward on the aggregated model to catch subtler anomalies that cannot be detected beforehand. By separating PSE and PAA, we provided both efficiency, and a second line of defense of additional protection, which enhances its robustness.

\subsection{Blockchain Components}
In this work, we implement a simple permissioned blockchain in which participants can join using a private key. An authentication server is considered to facilitate this task. Participants of this network are either users (i.e., formulated as $c_i\in C$) or minters (i.e., indicated as $c^*_i\in C^*$). Fig. \ref{fig:diagram} illustrates the design of \texttt{TrustChain} and shows how blockchain components are orchestrated along with the proposed algorithms.

\subsubsection{Blocks} Blocks $b^t\in B$ of the DFL contain commonly used header information such as hash to the previous block, timestamp $t$, and block number (i.e., indicated as in the subscript of $b$). Additionally, the header includes a flag that indicates whether this block contains $\theta_i^t$ or $\bar{\theta}^t$. In addition, a signature is included and used for authentication. Moreover, blocks contain data segments consists of $\bar{\theta}^t$, $\operatorname{HSIC}(\theta_m^{t-1}, \bar{\theta}^t)$, and $\{S_i^t\}_{i=1}^q$. Furthermore, blocks contain SCs that will be triggered when certain conditions are satisfied.

\subsubsection{Smart contracts} \texttt{TrustChain} uses three SCs that are triggered when these conditions are met:
\begin{enumerate}
    \item $SC_1$: Miners $c_i^*\in C^*$ request issuing $\bar{\theta}^{t+1}$ for $b^{t+1}$. This triggers PSE (Algorithm \ref{alg:selection}) to select $c_a$.
    \item $SC_2$: $c_a$ requests pushing $\bar{\theta}^{t+1}$ to $b^{t+1}$, which triggers PAA (Algorithm \ref{alg:validation}) to verify the issued update before adding it to $b^{t+1}$. If the condition is not met, $c_i$ corresponding to $c_a$ will not be selected for several rounds, and $SC_1$ is repeated.
    \item $SC_3$: If $\bar{\theta^{t+1}}$ is approved, the estimated $\operatorname{HSIC}(\theta_m^{t-1}, \bar{\theta}^t)$ in $SC_2$ will be included in $b^{t+1}$.
\end{enumerate}

\subsection{Time Complexity}
Let $d$ denote the model parameters' dimensionality, $k$ the number of candidate miners, $r$ the number of client updates in the current round, and $q$ the window size for historical statistics.

In PSE, the trust score of each candidate miner is computed as a decayed sum of cosine similarities over the last $q$ rounds. Computing a cosine similarity between two $d$-dimensional vectors requires a dot product and two norm calculations, each taking $O(d)$ operations, resulting in $O(d)$ per similarity. Since each of the $k$ candidates is evaluated over $q$ past rounds, PSE performs $kq$ such computations per round. Therefore, the per-round time complexity of PSE is
$O(k q d)$.

PAA begins by computing the elementwise median of $r$ client updates, each of dimension $d$. This can be achieved in $O(r d)$ using linear-time selection for each coordinate. It then evaluates the HSIC between the median update and the proposed aggregated model. For a generic kernel, HSIC computation involves constructing $d \times d$ Gram matrices and centering them, leading to $O(d^2)$ complexity. If a linear kernel is used, this reduces to $O(d)$. Finally, the threshold estimation step uses only $q$ stored HSIC values and requires $O(q)$, which is negligible compared to the other terms. Thus, the per-round complexity of PAA is
$O(r d + d^2 + q)$.

Combining both stages, the per-round time complexity of \texttt{TrustChain} is $O(k q d) + O(r d + d^2 + q) = O(k q d + r d + d^2)$. For high-dimensional models where $d \gg r$, the $O(d^2)$ HSIC term dominates. If a linear kernel or approximation is applied, the overall complexity reduces to $O(k q d+ rd)$, scaling linearly with the model size and number of candidates or participants.

\section{Experimental Results}
\label{sec:results}
This section first explains the experimental setup used in our experiments and then evaluates the proposed DFL under several scenarios. The experiments are conducted in three phases. First, we evaluate the overall accuracy when \texttt{TrustChain} is employed. Then, the sensitivity to attacks is discussed. Finally, we study PSE and PAA by isolating each component while using \texttt{TrustChain}.

\subsection{Experimental Setting}

\subsubsection{Datasets}
Experiments are evaluated using commonly used benchmark datasets from different domains, such as MNIST \cite{mnist}, Fashion-MNIST \cite{xiao2017fashionmnist}, CIFAR10 \cite{cifar}, Shakespeare \cite{pmlr-v54-mcmahan17a}, and Credit Card \cite{default_of_credit_card_clients_350} datasets. To create federated datasets, datasets are shuffled at first and then sub-datasets are created by random sampling without replacement. The resulting data is highly imbalanced, and the created subsets have an equal number of samples.

\subsubsection{Local models}

The architecture of local models differs for each dataset. For experiments on MNIST, Fashion-MNIST, and Credit Card a Multi-Layer Perceptron (MLP) with two hidden layers is used. For CIFAR10, on the other hand, we use a Convolutional Neural Network (CNN) with six convolutional layers followed by two dense layers. Convolution layers are paired with batch normalization, and after every two convolutional layers, max pooling and dropout (ratio=$25\%$) are used. In both networks, hidden layers use ReLU, and the final layer is activated using Softmax. Model parameters are optimized using stochastic gradient descent. The batch size is set to 64 for MLP and 32 for CNN. For the Shakespeare dataset, we use a Long Short-Term Memory (LSTM) network with two LSTM layers and a softmax activation.

\subsubsection{DFL parameters}
The DFL undergoes 1000 training iterations ($1\leq t\leq 1000$). The number of participants in $C$ is set to $r=100$. We assume $C^* \subset C$ and set $k=20$. The window length is set to $q=15$. The scaling factor in (\ref{eq:tau}) is empirically selected and set to $\lambda=0.5$. To facilitate the estimation of $\tau$, we assume the first 50 DFL iterations are safe, and attacks begin at $t=51$.

\subsubsection{Attacks scenarios}
Using label flipping, the local data on the attacker's side is poisoned with $\gamma=1$. Local training on the poisoned data results in malicious parameters $\tilde{\theta}$ which are sent to the DFL. Sign flipping is performed by training on clean local data and reversing the sign of gradients. GM is implemented by adding Gaussian noise to $\theta_i$ with a mean and standard deviation equal to $0.1$. For TOM, we set $\zeta=0.1$ in the corrupted loss function.

\subsubsection{Evaluation metrics}
The overall performance of the DFL structure is evaluated in terms of accuracy. Experiments are repeated ten times and the results are averaged. To evaluate PSE and PAA, we measure precision, as it takes both the success rate (true positives) and false alarms (false negatives) into account. 

\subsubsection{Computing infrastructure}
Results were obtained on a computer equipped with an NVIDIA RTX 3080 GPU, an Intel Core i7-12700 CPU, and 32 GB of RAM. The experiments were conducted using Python on an Ubuntu kernel, accessed via the Windows Subsystem for Linux.

\subsection{Results Analysis}
\begin{figure*}
    \centering
    \includegraphics[trim={0.8cm 0.4cm 0.4cm 0.4cm },clip,width=0.844\textwidth]{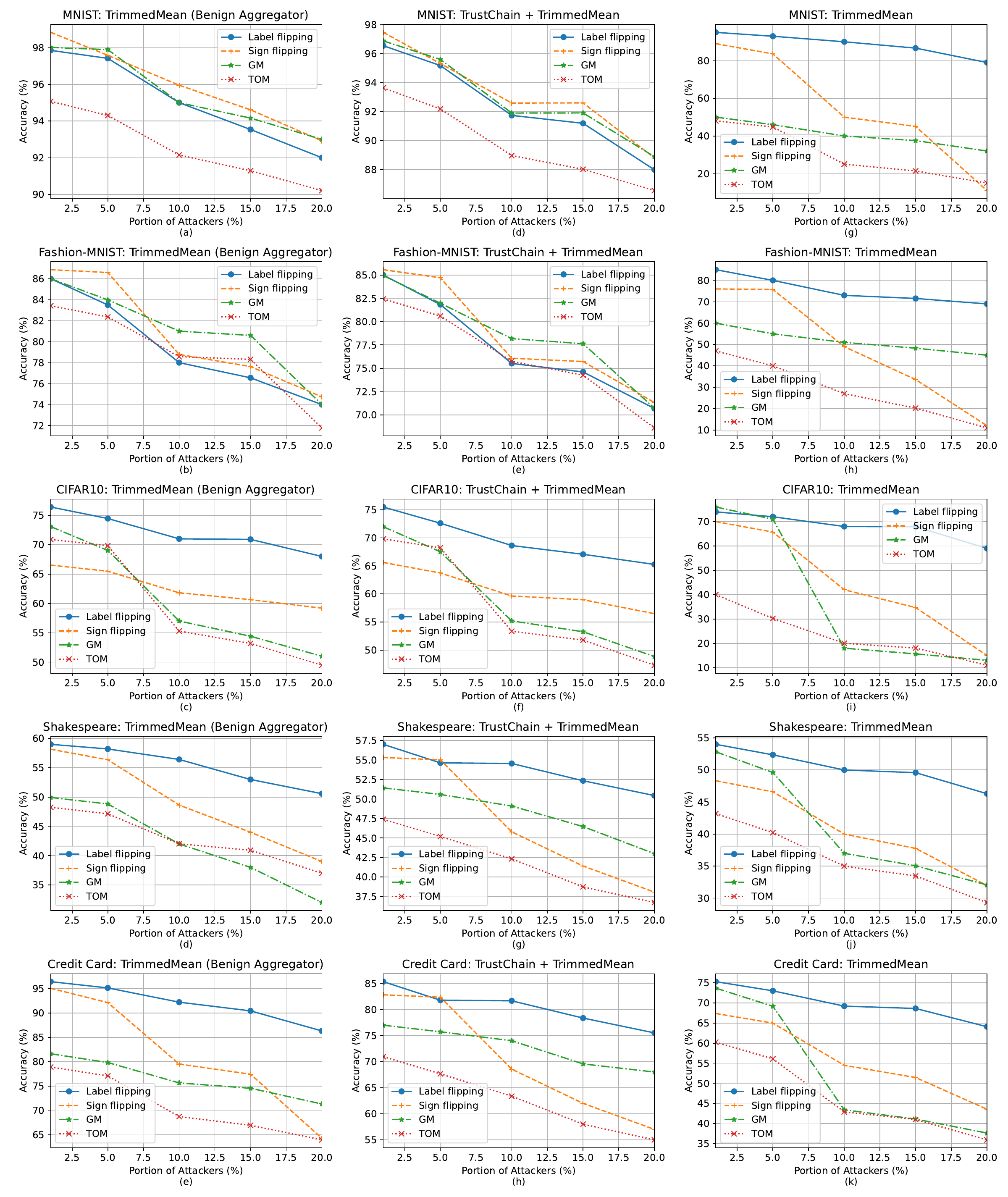}
    \caption{Classification accuracy of DFL for different datasets. On the left side, aggregator is benign. Plots in the middle column use \texttt{TrustChain} and aggregator is not trusted. On the right side, DFL do not use \texttt{TrustChain} and aggregator is not trusted.}
    \label{fig:attacks}
\end{figure*}

\subsubsection{Overall performance} In order to determine the effectiveness of \texttt{TrustChain}, we first compare the classification accuracy of DFL under four different attacks. In these experiments, malicious users attempt to inject poisonous updates into the DFL network. When DFL is not paired with \texttt{TrustChain}, aggregator nodes are selected randomly. Since the DFL is equipped with a robust aggregator, the effect of these attacks is expected to be partially mitigated when not initiated by the aggregator. Here, we select \texttt{TrimmedMean} \cite{pmlr-v80-yin18a} for the task of robust aggregation. Nevertheless, when these nodes are selected as the aggregator, they will have the opportunity to bypass the robust aggregator. Fig. \ref{fig:attacks} compares the classification accuracy of \texttt{TrimmedMean} under three conditions when poisoning attacks are launched with different ratios of Byzantine nodes, ranging from one to twenty percent of the nodes' population. To identify a baseline, we first investigate a scenario where the aggregator is benign and \texttt{TrustChain} is not used in the DFL, as shown in panels on the left side of Fig. \ref{fig:attacks}. This shows the performance of \texttt{TrimmedMean} in attack scenarios, without a malicious aggregator becoming involved. As expected, the accuracy deteriorates as the ratio of Byzantine nodes increases. Next, we assume the aggregator is not trusted, that is a malicious node can be chosen to carry out the aggregation. The middle panels in Fig. \ref{fig:attacks} shows the evaluation results of \texttt{TrustChain} under this condition. The accuracy obtained in these panels is slightly lower than that of the benign aggregator in Fig. \ref{fig:attacks}. This indicates that \texttt{TrustChain} can robustly mitigate malicious aggregators, albeit not completely. To reveal the effectiveness of \texttt{TrustChain} in this scenario, we remove it from the experiments, shown in plots on the right side of Fig. \ref{fig:attacks}, to demonstrate the potency of malicious aggregators when they are not audited. As illustrated, a noticeable performance drop has taken place as the bypassed poisoned update rapidly propagates through the DFL network. This effect becomes more pronounced as the ratio of Byzantine nodes increases because malicious aggregators will have a higher chance of being selected in each round.

\subsubsection{Sensitivity to attacks}

From Fig. \ref{fig:attacks}, it can be inferred that the portion of Byzantine nodes has a greater impact on their potency compared to the case when \texttt{TrustChain} is used. However, this observation is less noticeable for label flipping, as it varies within a smaller range. Moreover, TOM appears to be more difficult to mitigate, as it results in the lowest classification accuracy in most experiments shown in Fig. \ref{fig:attacks}. For large numbers of Byzantine nodes, sign flipping can also lead to catastrophic errors, when \texttt{TrustChain} is not employed. Nevertheless, the results indicate that \texttt{TrustChain} can mitigate the effect of the Byzantine population to a large extent. This can be seen by comparing plots in the middle and right side of Fig. \ref{fig:attacks}.

\subsubsection{Ablation study}
\begin{figure}
    \centering
    \includegraphics[trim={0.5cm, 0, 1cm, 0.9cm},clip,width=\columnwidth]{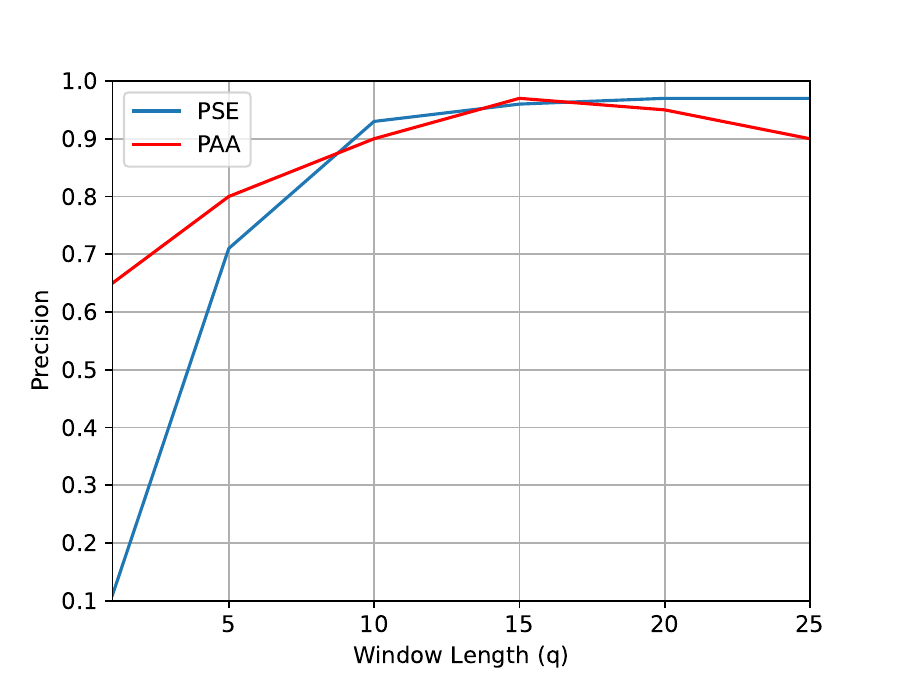}
    \caption{Effect of $q$ on the precision of PSE and PAA. Each component is studied separately, and the precision is averaged for all attack scenarios and datasets.}
    \label{fig:ablation}
\end{figure}

To study the efficacy of each component, we conduct two sets of experiments where only the selected component is exposed to adversaries. To test the PSE component, we launch the selected attacks from the participants' side and monitor the nodes PSE selects in this process. We estimate the precision by considering the selection of benign nodes as true positives and malicious nodes as false positives. Similarly, we use the same metric to evaluate PAA by considering correct and false verification as true positives and false positives, respectively. In this process, $\bar{\theta}$ is only corrupted by the aggregator, and nodes do not exhibit malicious behavior before being selected. Fig. \ref{fig:ablation} illustrates the precision recorded for PSE and PAA. The reported precision is averaged over the results of all attacks and datasets to capture the overall performance of these components. Based on this figure, it can be concluded that a small size of $q$ may lead to unreliable performance for both methods. For PSE, larger values of $q$ result in better precision because analyzing the drift over longer intervals is more accurate. However, the precision does not significantly change after a certain point. On the other hand, we notice that PAA is more sensitive to the choice of $q$. The plotted curve indicates that PAA works best for an optimal value of $q$, and values that are too small or too large result in deteriorated precision. This is due to the fact that when the window length is too small, PAA struggles to find an appropriate value for $\tau$. On the other hand, when $q$ is too large, older HSIC values that are too old are not removed from the window, leading to an inaccurate estimation of $\tau$, which only decreases and does not increase.

\begin{figure}
\centering

\includegraphics[trim={0.3cm 0.3cm 0.3cm 0.3cm},clip,width=\columnwidth]{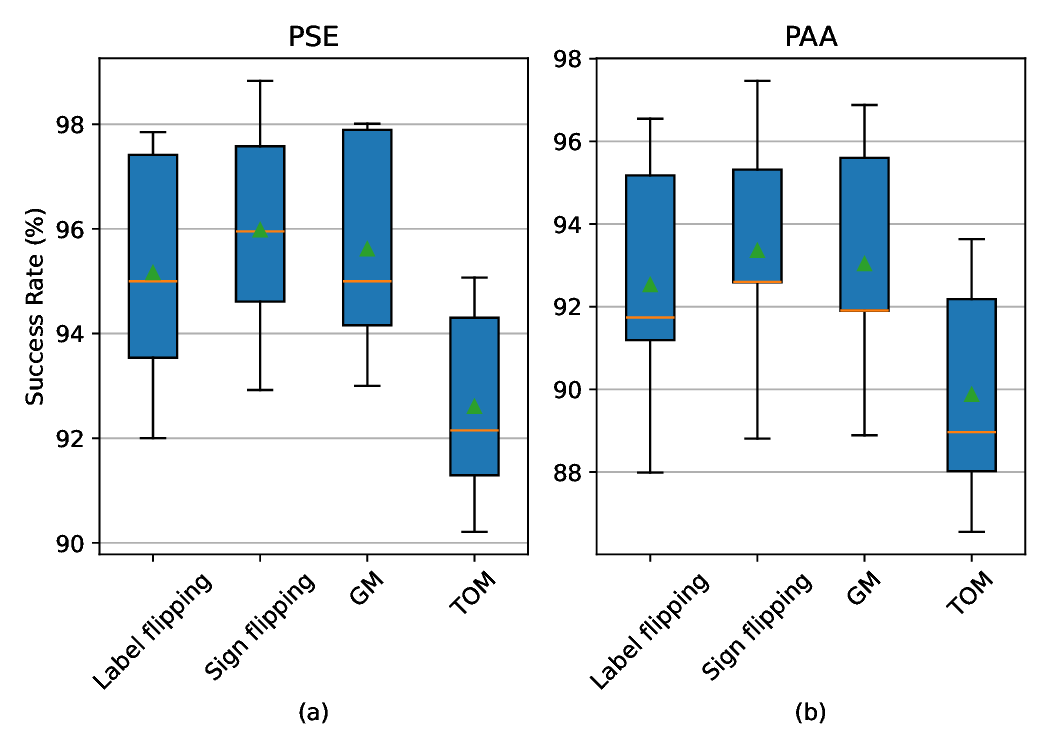}
\caption{Overall detection performance of PSE (used in SC1) and PAA (used SC2) for different attacks.}

\label{fig:boxplot}
\end{figure}

The overall success rates of PSE and PAA for each attack type are presented in Fig. \ref{fig:boxplot}. The consistently high detection rates of these two components demonstrate the effectiveness of SC1 and SC2 in mitigating malicious actors both before and after aggregator selection in each round. While the success rate remains high across all attacks, TOM proves more challenging to detect, resulting in a slightly lower success rate compared to the other attack types.

\subsubsection{Significance test}
\begin{figure}[t]
    \centering
    \includegraphics[width=\columnwidth]{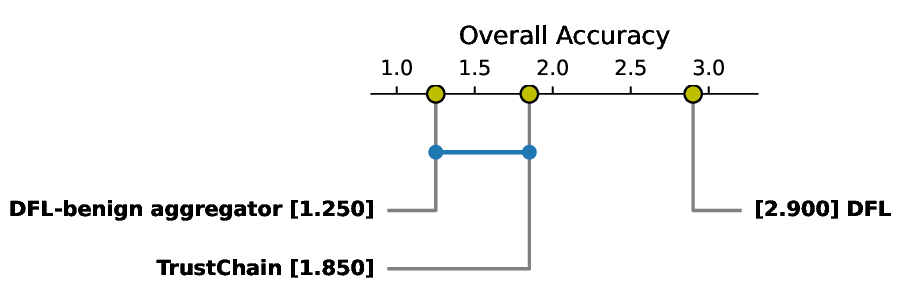}
    \caption{Critical difference diagram obtained from the post-hoc Friedman test. The significance level is set to $0.05$.}
    \label{fig:significant}
\end{figure}

Fig. \ref{fig:significant} shows Critical Difference (CD) diagrams obtained from the post-hoc Friedman test. The significance level (i.e., parameter $\alpha$) is set to $0.05$ in this test. This test estimates the significance of differences among the results obtained from each method across all experiments. Based on the determined CD level of this test, methods that are not significantly different in terms of accuracy are connected and grouped using colored lines. The illustrated CD diagram shows that employing \texttt{TrustChain} results in a performance that is statistically similar to that having a benign aggregator. Moreover, the proposed method significantly boosts the performance when the aggregator is not trusted (indicated as DFL in Fig. \ref{fig:significant}).

\section{Conclusion}
\label{sec:conclusion} 
This paper addressed the issue of malicious aggregators in DFL systems by proposing a blockchain-enabled solution for trustworthy aggregation, called \texttt{TrustChain}. The proposed method eliminates malicious aggregators in two steps. Firstly, the miners participating in the aggregation process are scored based on their tendency to drift from the distribution of aggregated parameters in previous DFL iterations. This step excludes miners who exhibit suspicious behavior. Secondly, the aggregated updates are checked before being pushed to the DFL by measuring statistical independence using HSIC between the aggregated model and the updates from this round. Aggregated models are approved based on a threshold that is dynamically adjusted according to the information available in previous blocks of the blockchain. This step adds an extra layer of security against trusted nodes that become rogue upon being selected as the aggregator. The proposed method was evaluated on three benchmark datasets and against four poisoning attacks, with varying numbers of Byzantine nodes. The results demonstrated the robustness of \texttt{TrustChain} in eliminating malicious aggregators. Future research directions will focus on studying \texttt{TrustChain} under more sophisticated attacks and integrating it with privacy-preserving methods to extend its applicability.


\bibliographystyle{IEEEtran}
\bibliography{main.bib}

\begin{IEEEbiography}[{\includegraphics[width=1in,height=1.25in,clip,keepaspectratio]{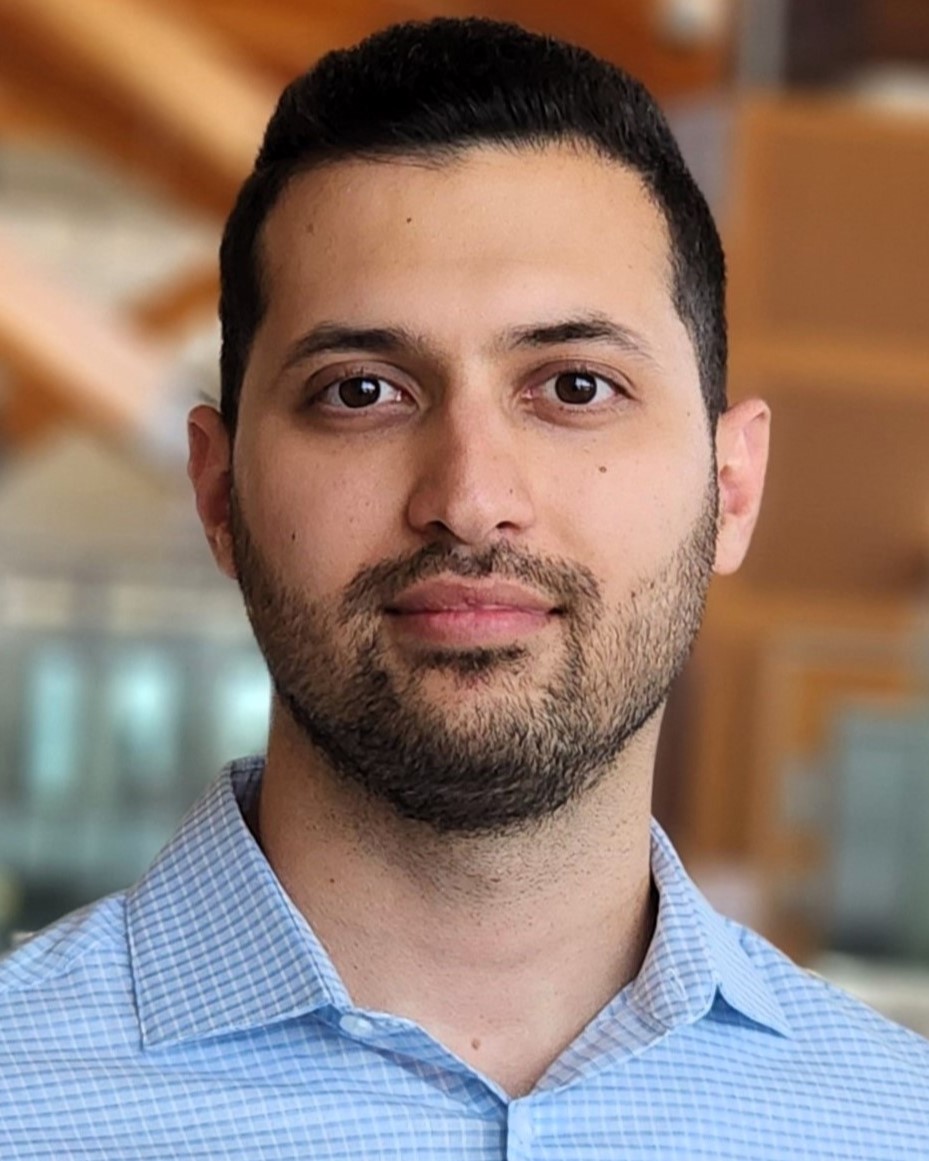}}]{Ehsan Hallaji}
(Member, IEEE) received the B.Sc. degree in software engineering from Shahid Rajaee University, Tehran, Iran, in 2015, and the M.A.Sc. and Ph.D. degrees in electrical and computer engineering from the University of Windsor, Windsor, ON, Canada, in 2018 and 2025, respectively. His doctoral research was recognized through prestigious national scholarships, including the NSERC Canada Graduate Scholarship (CGS-D3) and the Ontario Graduate Scholarship. He is currently a Machine Learning Engineer and his current research interests include machine learning, data mining, federated learning, and cybersecurity. He is a reviewer for several journals and conferences in his field of research. He also served as the Vice-Chair of the IEEE Systems, Man, and Cybernetics Society, Windsor Section, from 2019 to 2022.\\
\end{IEEEbiography}

\begin{IEEEbiography}[{\includegraphics[width=1in,height=1.25in,clip,keepaspectratio]{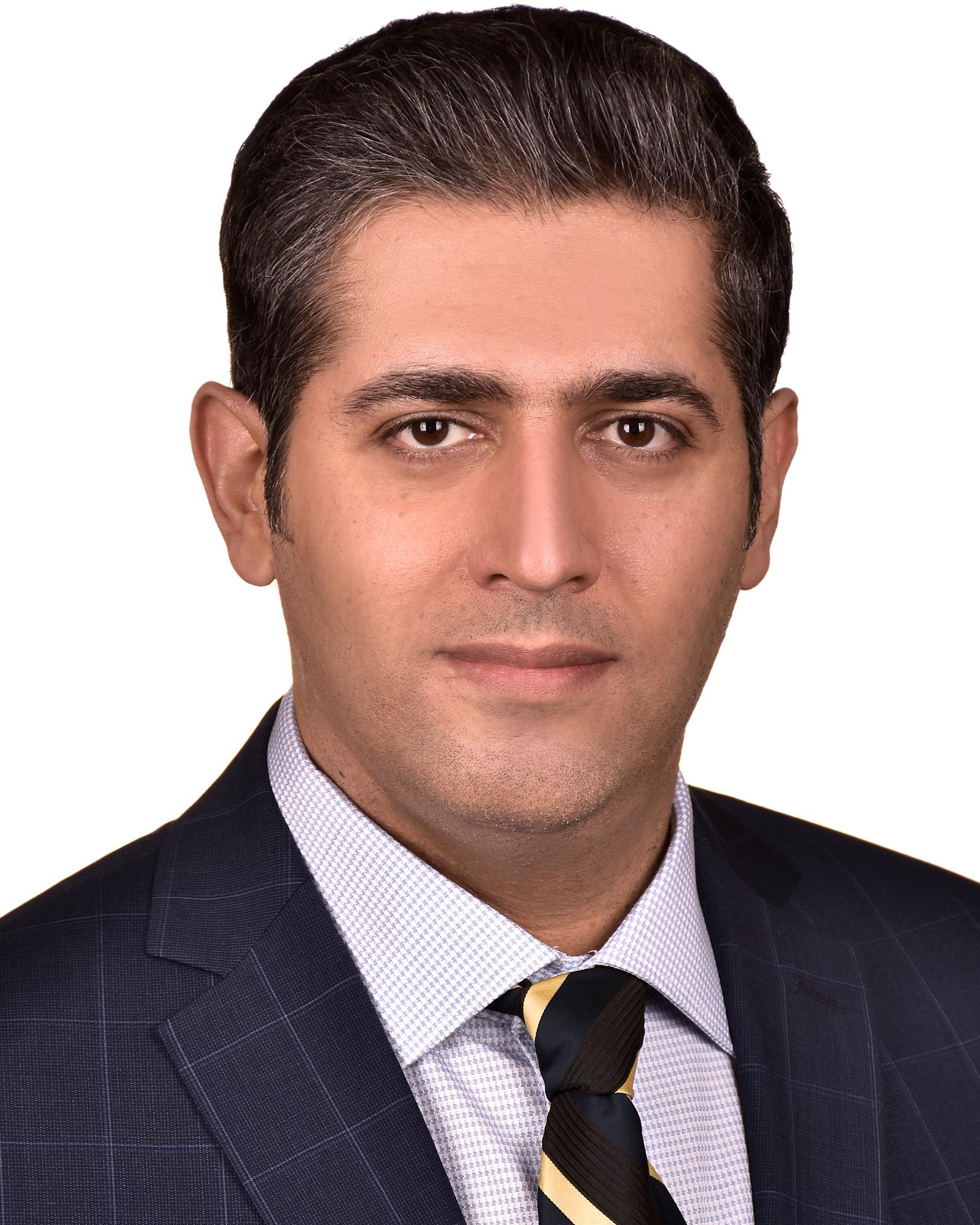}}]{Roozbeh Razavi-Far}
(Senior Member, IEEE) (Senior Member, IEEE) is an Associate Professor with the Faculty of Computer Science, University of New Brunswick and an Adjunct Research Professor in the Department of Computer Science at Western University, Canada. His research focuses on machine learning, adversarial machine learning, trustworthy and secure AI, big data analytics, computational intelligence, and cybersecurity of cyber-physical systems. He has authored or co-authored more than 180 papers in scholarly journals and international conferences. In 2024, Stanford listed him among the top two percent of most-cited researchers for the third consecutive year. He is the recipient of several awards and grants including NSERC-DG, NSERC-ECR, NBIF, USRG, MITACS, NCC R\&D, and NSERC-PDF. He is an associate editor at several journals, including the \textit{Neurocomputing}, \textit{Machine Learning and Knowledge Extraction}, \textit{Machine Learning with Applications}, \textit{Discover AI}, \textsc{IEEE Transactions on Industrial Cyber Physical Systems}, and \textsc{IEEE Access}. He served as a guest editor and chair for several journals and peer-reviewed conferences, and the chapter chair of IEEE Computational Intelligence, and Systems, Man and Cybernetics Societies at Windsor Section.\\
\end{IEEEbiography}

\begin{IEEEbiography}[{\includegraphics[width=1in,height=1.25in,clip,keepaspectratio]{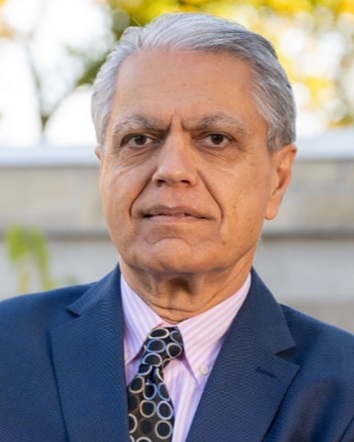}}]{Mehrdad Saif}(Fellow, IEEE) received bachelor’s, master’s, and doctoral degrees in electrical engineering, in 1982, 1984, and 1987, respectively. During his graduate studies, he was involved in research projects sponsored by NASA Lewis (now Glenn) Research Center, as well as the Cleveland Advanced Manufacturing Program.
He joined the School of Engineering Science, Simon Fraser University, Vancouver suburb, in 1987. He was a visiting scholar with the General Motors North American Operation Research and Development Center, NAO R\&D Center, Warren, MI, USA, from 1993 to 1994. From 2002 to 2011, he took on a senior academic leadership position by becoming the Director of the School of Engineering Science, SFU. In that capacity he oversaw a major expansion of that School. In 2004, he articulated the vision of starting the mechatronic systems engineering program at SFU’s newly established Surrey Campus. He planned and rolled out the program at both graduate and undergraduate levels in 2007.
In July 2011, Dr. Saif went on to take on the Dean of the Faculty of Engineering, University of Windsor, Windsor, ON, Canada. In his decanal role, he expanded the Faculty of Engineering programs into such areas as aerospace engineering, engineering management, Bachelor of Engineering Technology, mechatronics, and others. In his leadership roles at SFU and UWindsor, Dr. Saif’s efforts led to a significant enrollment increase at all levels, significant growth of the Faculty budget, modernization of the infrastructure, increased research productivity, as well as an increase in the number of faculty and staff supporting the engineering programs.
Dr. Saif is internationally recognized for his research work on the theory of state estimation, health monitoring, fault diagnosis, and prognosis, cybersecurity, attack detection, and attack resilience in complex engineering systems. Saif’s benchmark contributions have led to the advancement of estimation, control, machine learning, and AI theories, coupled with innovative developments and applications of said theories to a variety of cutting-edge real-world practical challenges in emerging areas of automotive, energy, biomedical, robotics, and aerospace industries. Dr. Saif is a Fellow of the IEEE and was cited for his contributions to monitoring, diagnosis, and prognosis in cyber-physical health systems. He is also a Fellow of Canadian Academy of Engineering, Fellow of the Engineering Institute of Canada, Fellow of the Institution of Engineering and Technology, Fellow of Asia-Pacific Artificial Intelligence Association, and a Fellow of Artificial Intelligence Industry Alliance.
Dr. Saif has served as a consultant to several industries and agencies, such as GM, NASA, BC Hydro, the Ontario Council of Graduate Studies, and others. He is the Editor-in-Chief of \textsc{IEEE Access} and serves on the editorial boards of several other IEEE and non-IEEE journals and conferences. Finally, he is also a registered Professional Engineer in the Province of Ontario.\\
\end{IEEEbiography}

\end{document}